\documentclass{article}
\usepackage{iclr2027_conference,times}

\usepackage{amsmath,amsfonts,bm}

\def\eqref#1{equation~\ref{#1}}

\def\1{\bm{1}}

\DeclareMathAlphabet{\mathsfit}{\encodingdefault}{\sfdefault}{m}{sl}
\SetMathAlphabet{\mathsfit}{bold}{\encodingdefault}{\sfdefault}{bx}{n}

\usepackage[utf8]{inputenc}
\usepackage[T1]{fontenc}
\usepackage{hyperref}
\hypersetup{hidelinks}
\usepackage{url}
\usepackage{graphicx}
\usepackage{booktabs}
\usepackage{amsfonts}
\usepackage{nicefrac}
\usepackage{microtype}
\usepackage{xcolor}
\usepackage{tcolorbox}
\usepackage{subcaption}
\usepackage{caption}
\usepackage{amsmath}
\usepackage{tabularx}
\usepackage{multirow}
\usepackage{colortbl}
\usepackage{wrapfig}
\usepackage{soul}
\usepackage{xspace}

\definecolor{sagegreen}{RGB}{87, 186, 173}
\definecolor{plannerblue}{RGB}{53, 83, 145}
\definecolor{reasonerorange}{RGB}{246, 173, 35}
\definecolor{generatorcyan}{RGB}{0, 174, 184}
\definecolor{criticred}{RGB}{194, 67, 21}
\definecolor{finalizergreen}{RGB}{93, 184, 18}

\newcommand{\planneragent}{\textcolor{plannerblue}{\textit{Planner Agent}}}
\newcommand{\reasoneragent}{\textcolor{reasonerorange}{\textit{Reasoner Agent}}}
\newcommand{\generatoragent}{\textcolor{generatorcyan}{\textit{Generator Agent}}}
\newcommand{\criticagent}{\textcolor{criticred}{\textit{Critic Agent}}}
\newcommand{\finalizeragent}{\textcolor{finalizergreen}{\textit{Finalizer Agent}}}

\newcommand{\ours}{\textit{REChart}\xspace}

\definecolor{rowgray}{gray}{0.95}

\title{REChart: Reasoning-Efficient Chart Editing with Large Reasoning Models}

\author{Yuanbang Liu, Chenxi Ruan, Yihan Hou, Qiong Luo \& Wei Zeng \thanks{ Wei Zeng is the corresponding author. E-mail: wei-zeng@hkust-gz.edu.cn.} \\
HKUST(GZ)\\
}

\iclrfinalcopy

\begin{document}

\maketitle

\begin{abstract}
  Chart editing requires inferring and modifying visualization code from a reference chart image based on an editing instruction, challenging fine-grained visual reasoning, instruction following, and executable code synthesis capabilities of MLLMs.
  Large reasoning models (LRMs) with extended Chain-of-Thought (CoT) reasoning are suitable for tackling such complex multimodal tasks.
  However, our preliminary study reveals an ``inverted-U'' relationship between reasoning length and chart-editing performance:
  Excessive reasoning often leads to ``overthinking,'' where models drift toward hallucinated visual details or get stuck in redundant reasoning loops.
  To address the gap, we introduce \ours, a two-stage training framework that provides process-level supervision over intermediate reasoning steps, improving both editing fidelity and reasoning efficiency.
  First, we synthesize 200k high-quality reasoning trajectories for supervised fine-tuning from a large image-instruction-code pool, using a role-specialized agentic Reason-Score-Refine workflow that iteratively refine the chart code toward higher quality.
  Second, we optimize the model via reinforcement learning with two complementary rewards: 
  a \emph{fidelity} reward evaluating code correctness, visual fidelity, and structural consistency, 
  and an \emph{efficiency} reward that assigns each rollout a random thinking budget, truncates the reasoning process, and credits the final reasoning segment according to its contribution to the output.
  On the ChartEdit and ChartMIMIC benchmarks, our model achieves state-of-the-art chart-editing performance among open-source models of comparable scale, while mitigating overthinking and reducing average reasoning token usage by 79.0\% under a maximum thinking budget of 16,384 tokens compared with the base model.
\end{abstract}

\section{Introduction}

Refinement of visual content plays an important role in scientific research and business intelligence~\citep{friendly2008brief}.
Chart editing, which involves iteratively adapting and refining visualization code by jointly interpreting source chart images and nuanced editing instructions, is a key aspect of this refinement process. 
Effective chart editing requires models to refine visual elements, adjust visual encodings, and adapt charts to new data, making it a challenging task for evaluating the reasoning and code generation capabilities of multimodal large language models (MLLMs)~\citep{yang2024chartmimic, zhao2025chartedit}.

The advent of large reasoning models (LRMs), characterized by extended Chain-of-Thought (CoT) processes, has shown immense potential in tackling complex reasoning tasks \citep{wei2022chain}.
While prior evidence suggests that scaling the length of reasoning tokens often correlates with improved performance \citep{guo2025deepseek, team2025kimi, jaech2024openai}, recent studies have identified a non-monotonic relationship \citep{aggarwal2025optimalthinkingbench, chen2025not, sui2025stop}.
Specifically, excessive reasoning may trigger overthinking, in which models become preoccupied with trivial details or amplify initial errors, leading to a noticeable decline in output quality \citep{feng2025what, wu2026when}.
To address this, several works have explored strategies such as length-penalized training, budget forcing, and adaptive reasoning depth to encourage more efficient use of reasoning tokens \citep{xiang2025just, muennighoff2025s1, li2025selfbudgeter}.
Despite these efforts, it remains unclear whether extended CoT improves chart editing, where precise visual perception must be coordinated with complex code synthesis.

We first conduct a preliminary study investigating the scaling behavior of LRMs on the ChartEdit benchmark~\citep{zhao2025chartedit} under varying thinking budgets.
Our findings reveal a distinct inverted-U relationship between reasoning token length and model performance: while initial CoT expansion enhances accuracy, excessive reasoning eventually triggers performance degradation as the model hallucinates visual details and falls into circular reasoning.
Besides, we observe that stronger models reach their peak performance with shorter reasoning chains, while harder tasks demand longer optimal budgets.
Together, these findings expose interrelated weaknesses of current LRMs in chart editing: 
\emph{reasoning efficiency} and \emph{editing fidelity}, as models routinely overshoot the optimal thinking budget that may lead to visual hallucinations and inconsistent code generation.
While recent advances in multimodal RL have introduced dense rewards for terminal code structures \citep{chen2025breaking, chen2025learning}, they largely leave the intermediate reasoning process unsupervised, which may lead to correct code via inconsistent thinking.

To enhance the model's reasoning efficiency and editing fidelity, we propose a two-stage training framework.
In the first stage, we construct a large-scale SFT dataset of 200k trajectories via a multi-agent collaborative generation framework.
Our framework generates trajectories that instantiate a Reason-Score-Refine loop, progressively refining intermediate outputs toward high-quality code generation.
In the second stage, we optimize the model via reinforcement learning with a hybrid reward that jointly guides the reasoning process and the final output.
Specifically, the reward combines an efficiency reward that samples a thinking budget for each rollout and credits the final reasoning segment before truncation, and a fidelity reward evaluating the finally generated chart across three dimensions, including code correctness, visual fidelity, and structural consistency.

We evaluate our model on ChartEdit and ChartMimic benchmarks across multiple tasks, achieving state-of-the-art performance among models of comparable scale. 
Compared to the base model, our approach yields a 13.3-point average overall performance gain across ChartEdit, ChartMIMIC Customized Mimic, and ChartMIMIC Direct Mimic.
Beyond accuracy, our method substantially improves reasoning efficiency.
Under a maximum thinking budget of 16,384 tokens, average reasoning token consumption is reduced by 79.0\% relative to the base model, with the largest reduction of 83.0\% on ChartMIMIC Customized Mimic.

In summary, our contributions are as follows:
\begin{itemize}
    \item We present a systematic empirical study revealing a distinct inverted-U relationship between reasoning token length and chart editing performance.
    \item We propose \ours, a two-stage training framework comprising SFT with a structured Reason-Score-Refine trajectory dataset and reinforcement learning with a hybrid reward guiding both the reasoning process and final output.
    \item Extensive evaluations demonstrate that \ours achieves state-of-the-art chart editing performance among open-source models of comparable scale, with a 13.3-point average overall improvement and 79.0\% reduction in reasoning token cost over the base model.

\end{itemize}

\section{Related Work}

\subsection{MLLMs for Chart Understanding}
Chart understanding with MLLMs is a rapidly growing field with many directions~\citep{huang2025pixel,davila2020chart}.
Some works focus on generating natural language outputs from charts, including chart question answering and summarization~\citep{masry2022chartqa, masry2023unichart, liu2024mmc, zeng2024advancing, rahman2023chartsumm, kantharaj2022chart, zhang2024tinychart}, and also table extraction from charts to facilitate reasoning~\citep{liu2023deplot, liu2023matcha, xu2024chartmoe, meng2024chartassistant}.
Other works focus on chart-to-code generation, with dedicated benchmarks covering diverse chart types and evaluation dimensions \citep{yang2024chartmimic, wu2025plot2code, zhao2025chartcoder, zhao2025chartedit}.
To improve generation fidelity, recent methods leverage reinforcement learning with both textual and visual feedback.
For example, \citet{tan2025chartmaster} proposed a chart similarity reward measuring both attribute and visual similarity, while \citet{chen2025breaking} introduced a multi-granularity structured reward combining textual and visual evaluation.
While these methods improve chart-to-code generation quality, they do not optimize the reasoning process itself, nor do they explore the applicability of their approaches to large reasoning models.
It remains unclear how to harness stronger reasoning capabilities efficiently for chart editing, a task that inherently demands long and complex reasoning to interpret visual elements, understand user intent, and synthesize precise code modifications.

\subsection{Large Reasoning Models}
Large reasoning models (LRMs) such as OpenAI o1~\citep{jaech2024openai}, DeepSeek-R1~\citep{guo2025deepseek}, and Kimi k1.5~\citep{team2025kimi1.5} extend the capabilities of language models by generating explicit chain-of-thought reasoning traces before producing a final answer.
This paradigm has demonstrated substantial gains on tasks~\citep{wei2022chain, sprague2024cot, chen2022program} such as complex logical inference, mathematical reasoning, and structured code generation, where deliberate step-by-step thinking proves critical to correctness.
However, the extended reasoning process also introduces significant token overhead, as models tend to generate verbose and sometimes redundant thinking traces regardless of task~\citep{sui2025stop,balachandran2025inference}.
This has motivated a growing line of work on reasoning efficiency, exploring strategies such as length-penalized training~\citep{xiang2025just, liu2025learn,aggarwal2025l1, hou2025thinkprune}, budget forcing~\citep{muennighoff2025s1, han2025token, yang2025dynamic}, and adaptive reasoning depth~\citep{li2025selfbudgeter, zhang2025adaptthink, xiang2025just, wang2025think} to reduce unnecessary computation while preserving output quality.
Chart editing is a particularly representative case, as it demands both fine-grained visual perception and structured code synthesis, making efficient reasoning critical for both quality and computational cost.

\section{Problem Definition and Preliminary Study}
\label{sec:preliminary_study}

\subsection{Problem Definition}
\label{sec:task_definition}
We focus on the chart-editing task, where a model generates updated chart code given a reference chart image and a natural-language editing instruction. 
A chart is formally represented as a tuple $(I, C)$, where $I$ denotes the rendered image and $C$ denotes the executable script (e.g., using Matplotlib or Seaborn in Python) that produces $I$. 
Given a reference chart image $I_{\text{ref}}$ and an editing instruction $T$, a model $\mathcal{M}$ outputs an output script:
\begin{equation}
C_{out} = \mathcal{M}(I_{\text{ref}}, T).
\end{equation}
The reference code $C_{\text{ref}}$ remains hidden during inference, requiring the model to ground edits in visual perception rather than direct code manipulation.
Our goal is to develop a model that can effectively grounds edits along three dimensions:
\emph{code correctness} and \emph{structural consistency} of $C_{out}$, and \emph{visual fidelity} of $I_{out}$ generated from $C_{out}$.  

Given the diversity of visual components and editing instructions, chart editing is inherently complex.
To systematically address this complexity, we built a taxonomy of edit operations through a multi-stage process. 
First, we aggregated a diverse set of editing samples from prior chart-editing datasets~\citep{zhao2025chartedit, chartm3} 
and editing tasks documented in the visualization community~\citep{dynavis, xieli_datawink}. 
Next, two visualization experts independently labeled and clustered these samples based on the induced operation. 
The experts focused on the type of modification (e.g., adding a data series, changing a color palette, rotating axis labels, filtering data points) rather than the specific code implementation. 
Through iterative consolidation, this process yielded six non-overlapping edit categories formally defined in Table~\ref{tab:task_definition}.
Among these, \emph{Direct Reproduction} serves as a no-edit baseline isolating raw image-to-code capability, and the remaining five categories encompass the diverse edit operations we distilled.


\begin{table}[t]
\centering
\caption{Taxonomy of chart-editing tasks across six functional categories.}
\label{tab:task_definition}
\small
\begin{tabularx}{\linewidth}{l l}
\toprule
\textbf{Category} & \textbf{Key Editing Aspects} \\
\midrule
Direct Reproduction   & No-edit baseline: reconstruction of the input chart from image to code. \\
Content Manipulation  & Data filtering, aggregation, point/series addition or removal, and trend-line overlays. \\
Style \& Aesthetics   & Modifications to backgrounds, grids, markers, colormaps, and visual effects. \\
Layout Edits          & Subplot composition, legend and colorbar positioning, spacing, and figure sizing. \\
Chart Transformation  & Chart type swapping, orientation changes, and coordinate system transformations. \\
Axes \& Annotations   & Title and label content and typography, axis scaling, ticks, and data annotations. \\
\bottomrule
\end{tabularx}
\end{table}

\subsection{Reasoning Scaling Behavior of LRMs}

\begin{figure*}[t]
\centering
\begin{subfigure}[t]{0.33\textwidth}
    \centering
    \includegraphics[width=\linewidth]{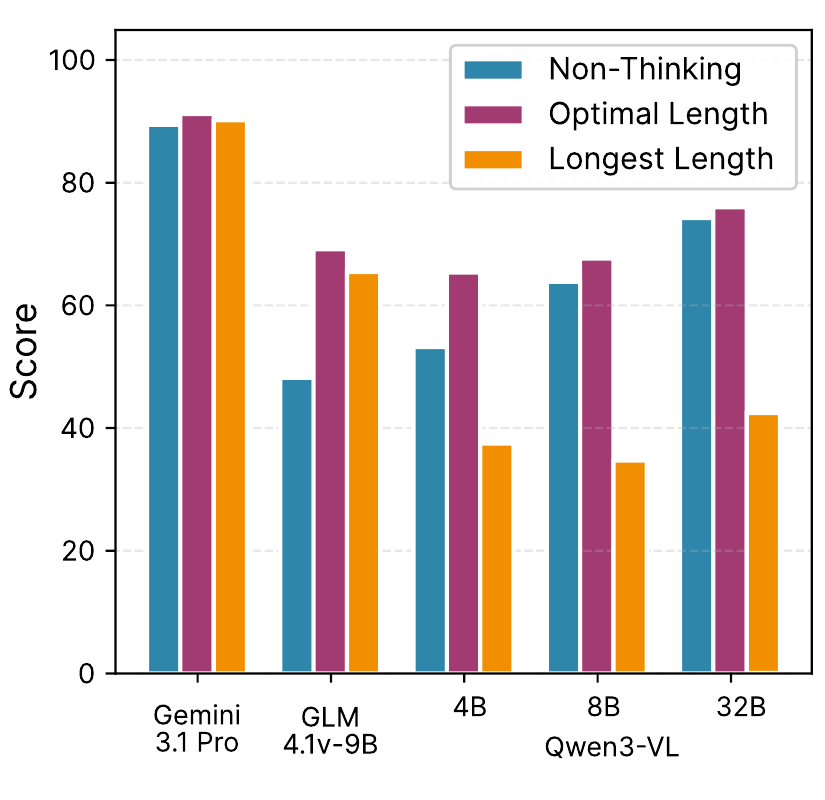}
    \caption{Model performance under non-thinking, optimal-budget, and longest-budget settings.}
    \label{fig:models_cot_len}
\end{subfigure}%
\hspace{0.1\textwidth}
\begin{subfigure}[t]{0.33\textwidth}
    \centering
    \includegraphics[width=\linewidth]{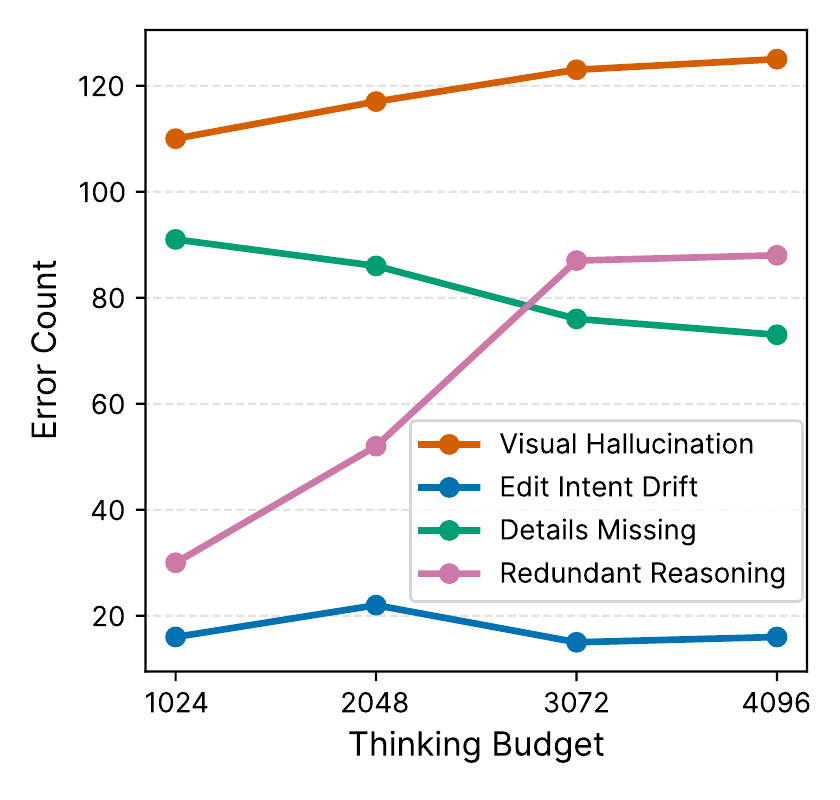}
    \caption{Error patterns in Qwen3-VL-8B reasoning under increasing thinking budgets.}
    \label{fig:budget_error_patterns}
\end{subfigure}
\caption{
Reasoning scaling on ChartEdit. (a) Performance follows an inverted-U trend with reasoning length. (b) Error patterns shift toward hallucination and redundancy as the budget grows.
}
\label{fig:preliminary_study}
\end{figure*}

We conduct an empirical study on the ChartEdit benchmark~\citep{zhao2025chartedit}, evaluating the Qwen3-VL series (4B, 8B, 32B)~\citep{bai2025qwen3}, GLM-4.1V-9B~\citep{hong2025glm}, and Gemini 3.1 Pro~\citep{gemini3} under varying thinking budgets.

Figure~\ref{fig:models_cot_len} shows a consistent inverted-U pattern between reasoning length and chart-editing performance.
Allowing LRMs to reason up to an appropriate budget improves accuracy over the non-thinking setting, indicating that explicit reasoning helps models coordinate visual perception, instruction following, and code synthesis.
However, extending the budget to the longest setting often degrades performance, especially for open-source models.
This suggests that additional reasoning is not uniformly beneficial: after a model passes its effective reasoning budget, the chain of thought can drift away from the visual evidence and reduce final execution or visual fidelity.

To understand why this degradation occurs, we further analyze Qwen3-VL-8B on the ChartEdit human set and categorize the errors appearing in its reasoning traces under different thinking budgets.
As shown in Figure~\ref{fig:budget_error_patterns}, longer budgets do not simply add useful intermediate steps.
Instead, visual hallucination steadily increases as the model spends more tokens describing chart elements that are not present or over-specifying visual details that later misguide the generated code.
Redundant reasoning also grows sharply with the budget, showing that the model often revisits the same observations without producing new evidence for the edit.
By contrast, details-missing errors decrease as the budget increases, indicating that longer reasoning initially helps the model inspect more visual components.
The resulting trade-off explains the inverted-U behavior: moderate reasoning reduces under-analysis, but excessive reasoning introduces hallucinated or repetitive intermediate claims that eventually harm chart-editing performance.

\section{REChart}
\label{sec:method}

\subsection{Reasoning Trajectory Construction for Chart Editing}
To train a model capable of efficient and high-fidelity chart-to-code generation, we build the supervised training data by first curating an image-instruction-code corpus from real-world scientific figures and then synthesizing structured reasoning trajectories from it.

\subsubsection{Instruction Corpus Construction}
\label{sec:instruction_corpus_construction}
\begin{figure*}[htb]
    \centering
    \includegraphics[width=\textwidth]{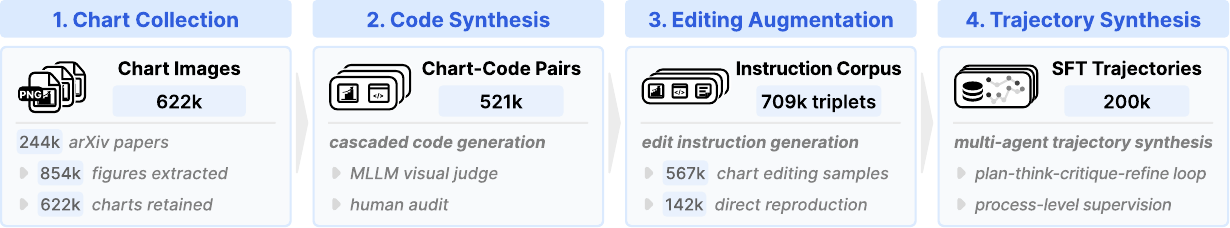} 
    \caption{The data collection and annotation pipeline consists of three stages: (1) Chart Collection and Filtering; (2) Chart-Code Pair Generation; and (3) Chart Editing Augmentation.}
    \label{fig:data_collection_pipeline}
\end{figure*}

In this work, we construct a real-world dataset of 709k image-instruction-code triplets for fine-tuning.
The corpus construction pipeline has three stages: 1) chart image collection, 2) code synthesis, and 3) editing augmentation (see Figure~\ref{fig:data_collection_pipeline}).

\textbf{Chart Image Collection.}
To cover diverse real-world chart styles and topics, we collect 2024 arXiv papers from computer science, bioinformatics, and mathematics, retaining roughly 100k CC BY 4.0 articles.
From their \LaTeX{} sources, we extract 854k embedded figures and use an MLLM classifier to filter out non-chart figures, yielding 622k chart images.

\textbf{Code Synthesis.}
To balance efficiency and quality, we synthesize visualization code with a tiered set of generators (Qwen3-VL-32B, Qwen3-VL-235B-A22B, and Gemini~3~Flash).
Rather than requiring pixel-level reproduction, we retain image-code pairs that pass a Qwen3-VL-32B audit on readability, completeness, visual integrity, and code-visual consistency; failed cases are forwarded to the next generator in the cascade.
To verify audit reliability, we manually inspect a random sample of 1,000 pairs.
This process yields 521k high-quality image-code pairs, with full prompts and filtering statistics provided in Appendix~\ref{app:data_construction_details}.

\textbf{Editing Augmentation.}
We then build the editing portion of the dataset by prompting Qwen3-VL-32B to generate instruction-code pairs over the five editing categories defined in Table~\ref{tab:task_definition}.
Since each of the 521k chart-code pairs is prompted three times, the accepted candidates across all pairs amount to 567k editing samples, while the 142k pairs for which no candidate passes the audit are repurposed as direct-reproduction samples.
The final dataset comprises approximately 709k image-instruction-code triplets spanning the six functional dimensions of the taxonomy, covering both direct reproduction and chart editing tasks.
Detailed \mbox{pipeline statistics, model assignments, and prompts are provided in Appendix~\ref{app:data_construction_details}.}

\subsubsection{Reasoning Trajectory Synthesis}
\label{sec:trajectory_synthesis}

\begin{figure*}[htb]
  \centering
    \includegraphics[width=\linewidth]{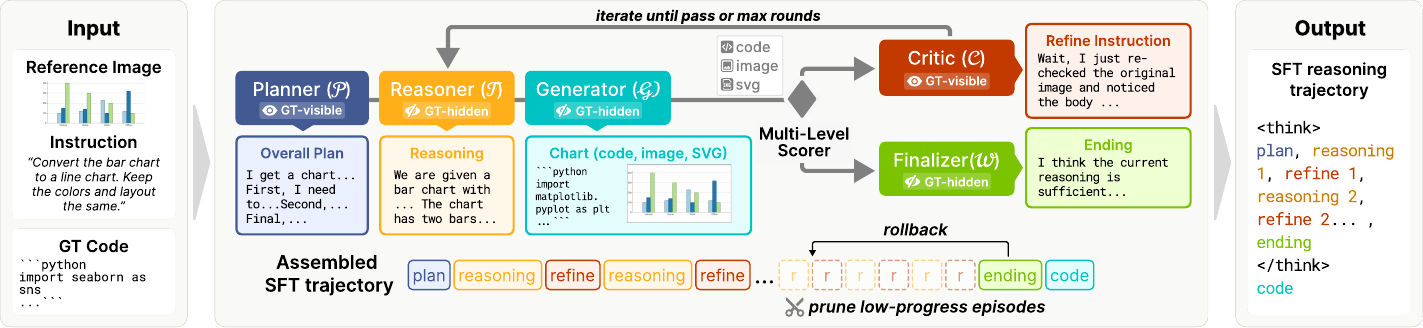}
    \caption{Multi-agent pipeline for generating the reasoning path.}
    \label{fig:multi_agent_pipeline}
\end{figure*}

Building on the preliminary study in Section~\ref{sec:preliminary_study}, we synthesize SFT trajectories that provide explicit step-level supervision to address the overthinking problem of LRMs in chart editing.
Naively prompting a single model to produce longer CoT traces does not resolve these issues, since the resulting traces are still optimized only by the final answer and may contain redundant or misleading reasoning steps.
To address these issues at the data-construction stage, we use a role-specialized multi-agent workflow as illustrated in Figure~\ref{fig:multi_agent_pipeline}.
Inspired by MRT that estimate single reasoning episode's contribution to the final outcome to guide the reasoning process~\citep{qu2025optimizing}, the workflow uses an external scorer to measure episode-level progress, prune stagnant reasoning segments, and redirect the trajectory through critique when the current path stops improving.

\textbf{(1) Agent Roles.}
The system comprises five role-specialized agents, each operating under a dedicated prompt.
The \planneragent{} ($\mathcal{P}$) receives the reference image, the editing instruction, and the ground-truth code, and produces a concise overall plan that decomposes the task into a sequence of high-level steps.
The \reasoneragent{} ($\mathcal{T}$) operates only on the reference image, the instruction, and the overall plan, and at each iteration appends a single new reasoning episode to the running context, mimicking the step-wise self-talk of o1-style reasoning.
The \generatoragent{} ($\mathcal{G}$) consumes the accumulated context and emits a candidate Python script.
The \criticagent{} ($\mathcal{C}$) is invoked when a candidate fails verification, and produces a first-person refine instruction (e.g., ``Wait, I should re-examine the legend layout$\ldots$'') that pinpoints a concrete weakness in the current reasoning, framed as the model's own self-correction inside the trajectory.
The \finalizeragent{} ($\mathcal{F}$) is invoked once a candidate clears the quality bar, and composes a brief concluding thought that closes off the chain of thought before the final code block.
The detailed implementation and setups of the pipeline are detailed in Appendix~\ref{app:agent_prompts}.

\textbf{(2) Episode-Level Reason-Score-Refine Loop.}
A trajectory begins with $\mathcal{P}$ producing the overall plan, after which the system enters an iterative loop in which each iteration constitutes one episode.
At iteration $t$, $\mathcal{T}$ extends the context with a new reasoning episode, $\mathcal{G}$ produces candidate code from the updated context, and an external multi-level scorer $s_t \in [0, 100]$ evaluates the rendered chart against the reference image using the instruction, the code, and an intermediate SVG (see in Section~\ref{sec:fidelity_reward}).
We treat the score change $\Delta_t = s_t - \max_{t' < t}\, s_{t'}$ as a measurable proxy for the per-episode \emph{progress} in the sense of~\citet{qu2025optimizing}: an episode that advances the running best contributes to eventual success, while one that fails to do so accrues regret without informational return.

If $s_t$ reaches the acceptance threshold $\tau = 85$, the loop terminates and control passes to $\mathcal{F}$.
Otherwise, the system applies a stagnation check before continuing.
When three consecutive episodes have failed to surpass the running best ($\Delta_{t-2}, \Delta_{t-1}, \Delta_t \le 0$), we interpret this as a high-regret excursion: the corresponding three episodes are removed from the context, and the context is rolled back to the snapshot that produced the running best.
This explicit pruning prevents the trajectory from drifting along an unproductive path and bounds the regret accumulated within any single trajectory, which is the property our downstream SFT loss inherits from the data.
$\mathcal{C}$ then generates a refine instruction conditioned on the current state, which is appended to the context to bias the next episode toward the deficiency it identifies, and control returns to $\mathcal{T}$.
The loop is capped at 9 iterations, and trajectories that fail to converge within this budget are discarded.

\textbf{(3) Trajectory Assembly.}
The accepted trajectory comprises the overall plan from $\mathcal{P}$, an alternating sequence of reasoning episodes from $\mathcal{T}$ and refine instructions from $\mathcal{C}$, the closing thought from $\mathcal{F}$, and the final code from $\mathcal{G}$.
Following the standard o1-style format, the reasoning portion is wrapped in \texttt{<think>}\,$\ldots$\,\texttt{</think>} tags and followed by the final output with executable code inside.
By construction, the synthesized trajectories retain only reasoning episodes that collectively advance the fidelity score over successive ones, exposing the model to structured reasoning behavior where each retained sequence of steps contributes measurable progress toward the final output.

\subsection{Reinforcement Finetuning}
\label{sec:rlvr}

\begin{figure*}[htb]
  \centering
    \includegraphics[width=0.75\linewidth]{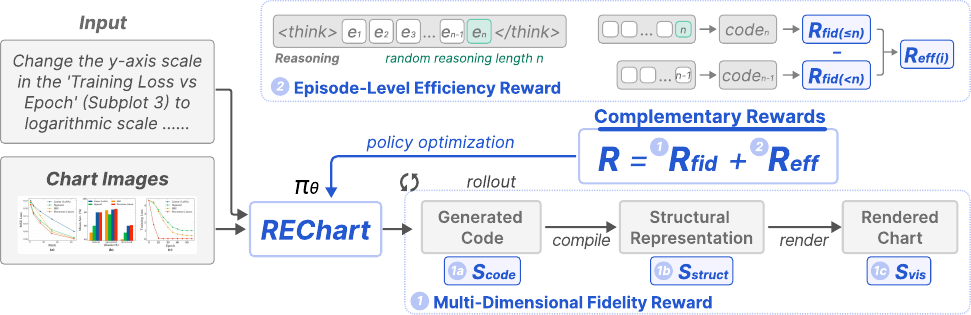}
    \caption{Overview of our RL framework guided by an episode-level efficiency reward and a multi-dimensional fidelity reward.}
    \label{fig:rl}
\end{figure*}
While SFT equips the model with structured reasoning behavior over synthesized trajectories, it may overfit to the synthetic data distribution and exhibit performance degradation on out-of-distribution editing instructions.
To address this, we further optimize the model via reinforcement learning with two complementary rewards: a fidelity reward evaluating the final generated chart across code correctness, visual fidelity, and structural consistency, and an efficiency reward that randomly budgets each rollout, truncates the reasoning process, and credits the final reasoning segment before the answer (see Figure \ref{fig:rl}).

\subsubsection{Fidelity Reward}

\label{sec:fidelity_reward}
For each rollout, the code extracted from the model's final output is executed in a sandboxed environment to obtain both the rendered SVG and a rasterized image. We assess fidelity across the following three dimensions.

\paragraph{Code Similarity.}
We apply CodeBLEU~\citep{codebleu} to measure the overall code similarity between the generated and ground-truth solution.

\paragraph{Structural Similarity.}
Rather than comparing raw SVG markup, we construct a lightweight structural representation of the chart.
Specifically, we parse the SVG into its DOM tree, prune rendering-irrelevant nodes, and classify the \texttt{path} elements into geometric primitives based on their \texttt{d} attributes.
Each node is then reduced to its essential attributes, including element type, text content, and color.
The resulting simplified trees of the generated and ground-truth charts are compared to produce a structural similarity score, reflecting \mbox{whether the chart's compositional layout is correctly reproduced.}

\paragraph{Visual Fidelity.}
We employ a multimodal embedding model Qwen3-VL-Embedding~\citep{li2026qwen3} to evaluate instruction-following quality.
Specifically, we measure the embedding similarity between the generated chart and the input context, which consists of the editing instruction paired with the reference chart image.
This score captures \mbox{whether the output faithfully reflects the required visual changes.}

The fidelity reward $R_{\text{fidelity}}$ is computed as a weighted average of the three dimension scores:
\begin{equation}
R_{\text{fid}} = w_{code} \cdot s_{\text{code}} + w_{struct} \cdot s_{\text{struct}} + w_{vis} \cdot s_{\text{vis}}
\end{equation}

\subsubsection{Efficiency Reward.}
Optimizing solely on the outcome reward (fidelity reward $R_{fidelity}$) suffers from the credit-assignment problem~\citep{qu2025optimizing}, where a successful rollout reinforces all intermediate reasoning episodes equally, regardless of their actual contribution to the final outcome.
Following~\citet{qu2025optimizing}, we incorporate an episode-level efficiency reward that measures whether the final reasoning segment before the answer improves the fidelity reward under a sampled compute budget.


Concretely, for each rollout, we randomly sample a thinking budget $B$ and allow the model to reason until it either emits the end-of-thinking token or reaches the budget.
If the rollout reaches $B$ before naturally stopping, we truncate the reasoning context and force the model to enter the answer phase.
Let $n$ denote the actual number of reasoning segments exposed under this procedure, which may be smaller than or equal to the sampled budget $B$.
We denote by $e_n$ the final reasoning segment immediately preceding the answer, by $h_{<n}$ the truncated context before $e_n$, and by $h_{\leq n}=h_{<n}\cup\{e_n\}$ the context that includes it.
We decode answers from both contexts, evaluate them under the fidelity reward to obtain $R_{\text{fid}}^{(<n)}$ and $R_{\text{fid}}^{(\leq n)}$, and define the efficiency reward as:
\begin{equation}
R_{\text{eff}}(e_n) = \max(R_{\text{fid}}^{(\leq n)} - R_{\text{fid}}^{(<n)}, 0).
\end{equation}
The reward therefore credits only the last reasoning segment actually exposed before answer generation, encouraging the model to make the reasoning immediately before generation concise and useful rather than rewarding arbitrarily selected spans from a completed trajectory.

\subsubsection{Policy Optimization}

The composite reward used for policy optimization is then:
\begin{equation}
R = w_{\text{fid}}\, R_{\text{fid}} + w_{\text{eff}}\, R_{\text{eff}}.
\end{equation}
We optimize the model with this reward via GRPO~\citep{shao2024deepseekmath}, which samples a group of candidate responses per input and normalizes rewards within the group to compute relative advantages for policy updates.

\begin{table*}[t]
\centering
\caption{Performance on ChartEdit \textit{w/o} Code and ChartMIMIC (customized mimic) benchmarks.}
\scriptsize
\label{tab:overall_results}
\setlength{\tabcolsep}{3pt}
\resizebox{\linewidth}{!}{
\begin{tabular}{l r | cccc | cccc}
\toprule
\multirow{2}{*}{Model} & \multirow{2}{*}{\# Params.} &
\multicolumn{4}{c|}{ChartEdit \textit{w/o} Code} &
\multicolumn{4}{c}{ChartMIMIC Customized} \\
\cmidrule(lr){3-6} \cmidrule(lr){7-10}
& & Exec. & Code & Chart & Overall &
  Exec. & Low & High & Overall \\
\midrule
\multicolumn{10}{c}{\textit{Proprietary Models}} \\
\midrule
GPT-4o~\citep{gpt4o} & - & 91.5 & 60.0 & 79.9 & 70.0 & 96.5 & 82.1 & 84.3 & 83.2 \\
Claude-3.7-Sonnet~\citep{claude37} & - & 96.3 & 73.4 & 87.5 & 80.5 & 80.6 & 67.4 & 78.1 & 72.3 \\
Gemini-2.5-Pro~\citep{gemini2_5} & - & 95.3 & 82.9 & 89.2 & 86.1 & 91.7 & 81.3 & 83.5 & 82.4 \\

\midrule
\multicolumn{10}{c}{\textit{Open-Source Models}} \\
\midrule
Qwen2.5-VL-72B~\citep{Qwen2.5-VL} & 72B & 89.8 & 57.7 & 71.0 & 64.4 & 85.3 & 67.6 & 69.1 & 68.4 \\
Qwen2.5-VL-32B~\citep{Qwen2.5-VL} & 32B & 88.4 & 61.7 & 72.5 & 67.1 & 85.6 & 66.4 & 69.5 & 68.0 \\
Qwen3-VL-32B-Thinking~\citep{bai2025qwen3} & 32B & 85.5 & 72.9 & 76.3 & 74.6 & 84.0 & 67.0 & 72.3 & 69.6 \\
Qwen3-VL-32B-Instruct~\citep{bai2025qwen3} & 32B & 84.9 & 69.0 & 73.4 & 71.2 & 88.2 & 72.4 & 61.3 & 66.9 \\
\midrule
Qwen3-VL-8B-Thinking~\citep{bai2025qwen3} & 8B & 79.1 & 66.6 & 68.0 & 67.3 & 77.6 & 59.6 & 63.4 & 61.5 \\
GLM4.1v-9B-Thinking~\citep{hong2025glm} & 9B & 78.9 & \underline{68.4} & 68.7 & 68.6 & 80.7 & \underline{63.7} & \underline{65.7} & \underline{64.7} \\
Qwen3-VL-8B-Instruct~\citep{bai2025qwen3} & 8B & 84.9 & 59.6 & 69.3 & 64.5 & 78.1 & 61.4 & 60.8 & 61.1 \\
LLaVA-1.5-13B~\citep{llava} & 13B & 48.8 & 11.4 & 16.7 & 14.1 & 49.0 & 23.6 & 24.7 & 24.2 \\
ChartLlama~\citep{chartllama} & 13B & 52.3 & 15.8 & 27.4 & 21.6 & - & - & - & - \\
ChartCoder~\citep{zhao2025chartcoder} & 7B & 89.8 & 47.3 & 74.5 & 60.9 & 85.2 & 58.3 & 62.0 & 60.2 \\
MSRL~\citep{chen2025breaking} & 7B & \textbf{97.2} & 57.6 & \underline{84.8} & \underline{71.2} & 67.1 & 39.6 & 46.5 & 43.0 \\
RRVF~\citep{chen2025learning} & 7B & 96.1 & 44.1 & 74.2 & 59.2 & \underline{85.9} & 57.6 & 59.3 & 58.5 \\
Qwen3-VL-4B-Instruct~\citep{bai2025qwen3} & 4B & 79.5 & 55.6 & 63.2 & 59.4 & 66.0 & 47.6 & 49.1 & 48.4 \\
Qwen3-VL-4B-Thinking~\citep{bai2025qwen3} & 4B & 79.3 & 64.3 & 67.0 & 65.6 & 72.9 & 54.1 & 57.6 & 55.8 \\
TinyChart~\citep{zhang2024tinychart} & 3B & 36.3 & 18.3 & 25.1 & 21.7 & - & - & - & - \\
ChartEditor~\citep{chen2026charteditor} & 3B & 66.5 & 40.2 & 55.3 & 47.8 & 61.6 & 52.1 & 57.9 & 55.0 \\
\midrule
\rowcolor{rowgray}
Ours & 8B & \underline{96.7} & \textbf{73.3} & \textbf{86.1} & \textbf{79.7} & \textbf{88.6} & \textbf{67.7} & \textbf{72.5} & \textbf{70.1} \\
\bottomrule
\end{tabular}
}

\end{table*}

\section{Experiment}
\label{sec:experiments}
\subsection{Baselines and Benchmarks}
\label{sec:baselines_benchmarks}
We compare \ours against three groups of baselines: proprietary MLLMs including GPT-4o~\citep{gpt4o}, Claude-3.7-Sonnet~\citep{claude37}, and Gemini-2.5-Pro~\citep{gemini2_5} as upper-bound references; general-purpose open-source MLLMs of comparable or larger scale, including the LLaVA-1.5-13B~\citep{llava}, the Qwen3-VL series~\citep{bai2025qwen3} and GLM-4.1V-9B-Thinking~\citep{hong2025glm}; and domain-specialized models including ChartCoder~\citep{zhao2025chartcoder}, ChartEditor~\citep{chen2026charteditor}, TinyChart~\citep{zhang2024tinychart}, and RRVF~\citep{chen2025learning}.
We evaluate on two benchmarks covering both chart editing and direct chart-to-code generation: ChartEdit \textit{w/o} code~\citep{zhao2025chartedit}, ChartMIMIC~\citep{yang2024chartmimic} (customized mimic and direct mimic).
Further details on evaluation configurations are provided in Appendix~\ref{app:eval_config}.
\subsection{Training Setup}

We fine-tune Qwen3-VL-8B-Thinking~\citep{bai2025qwen3} as our base model across two stages. In the SFT stage, we train on 200k synthesized trajectories for one epoch using LLaMA-Factory~\citep{zheng2024llamafactory}, with a learning rate of 5e-5 and an effective batch size of 256 via gradient accumulation, completing in approximately 28 hours on 8 NVIDIA A800 80GB GPUs. In the RL stage, we further train on 82k samples for one epoch using Verl~\citep{verl}, with a GRPO group size of 8. For the hybrid reward, the three fidelity dimensions are weighted equally ($w_{\text{code}} = w_{\text{struct}} = w_{\text{vis}} = 0.33$), and the fidelity and efficiency rewards contribute equally to the final reward ($w_{\text{fid}} = w_{\text{eff}} = 0.5$). This stage requires approximately 65 hours on 8 NVIDIA H100 GPUs.
\subsection{Main Results}
\label{sec:main_results}

\paragraph{Chart Editing.}

Table~\ref{tab:overall_results} reports results on ChartEdit \textit{w/o} Code and ChartMIMIC Customized Mimic set.
On ChartEdit, \ours achieves an overall score of 79.71, surpassing all open-source baselines, including models with substantially more parameters, such as Qwen3-VL-32B-Thinking (74.6) and Qwen3-VL-32B-Instruct (71.2).
The gains are consistent across all evaluation dimensions, with \ours attaining a code score of 73.28 and a chart score of 86.14, demonstrating improvements in both code-level correctness and chart-level fidelity.
On ChartMIMIC Customized Mimic, \ours achieves an overall score of 70.09, outperforming all open-source baselines, including Qwen3-VL-32B-Thinking (69.6), GLM4.1V-9B-Thinking (64.7), and Qwen3-VL-8B-Thinking (61.5).
Compared to domain-specialized baselines, \ours consistently outperforms prior chart-specific models such as ChartCoder, ChartEditor, and RRVF across both benchmarks.
While proprietary models such as Gemini-2.5-Pro retain an overall advantage, \ours substantially narrows this gap as an 8B open-source model.

\begin{wraptable}{r}{0.65\textwidth}
\vspace{-15pt}
\setlength{\tabcolsep}{3pt}
\caption{Results on ChartMimic Direct Mimic task.
}
\label{tab:main_results_direct_mimic}
\setlength{\tabcolsep}{2pt} 
\renewcommand{\arraystretch}{1.05}
\begin{small}
\begin{tabular}{l r c c c c}
\toprule
Model & Exec. & Low & High & Overall \\ 
\midrule
GPT-4o~\citep{gpt4o} & \textbf{93.2} & \textbf{79.0} & \textbf{83.5} & \textbf{81.2} \\
GeminiProVision~\citep{geminiprovision} & 68.2 & 53.8 & 53.3 & 53.6 \\
Claude-3-opus~\citep{Claude3opus} & 83.3 & 60.5 & 60.1 & 60.3 \\

\midrule
InternVL2-8B~\citep{internvl2series} & 61.8 & 34.4 & 38.9 & 36.6 \\
LLaVA-Mistral-7B~\citep{llava} & 59.7 & 20.7 & 21.3 & 21.0 \\
Qwen3-VL-32B-Thinking~\citep{bai2025qwen3} & 82.8 & 65.2 & 72.2 & 68.7 \\
Qwen3-VL-8B-Thinking~\citep{bai2025qwen3} & 75.0 & 56.0 & 60.9 & 58.4 \\
Qwen3-VL-4B-Thinking~\citep{bai2025qwen3} & 69.8 & 51.5 & 56.0 & 53.7 \\
GLM4.1v-9B-Thinking~\citep{hong2025glm} & 79.5 & 63.7 & 67.7 & 65.7 \\
ChartEditor~\citep{chen2026charteditor} & 69.6 & 52.7 & 56.4 & 54.5 \\
ChartCoder~\citep{zhao2025chartcoder} & {91.4} & \underline{72.5} & \underline{74.0} & \underline{73.3} \\
RRVF~\citep{chen2025learning} & \textbf{97.8} & 60.9 & 67.9 & 64.4 \\
Ours & \underline{94.7} & \textbf{72.8} & \textbf{81.5} & \textbf{77.2} \\
\bottomrule
\end{tabular}
\end{small}
\end{wraptable}

\paragraph{Direct Chart Reproduction.}
Table~\ref{tab:main_results_direct_mimic} reports results on the ChartMIMIC Direct Mimic set, which evaluates chart-to-code reproduction without any editing instruction.
\ours achieves an overall score of 77.15, outperforming all open-source baselines of comparable or larger scale, including ChartCoder (73.3), Qwen3-VL-32B-Thinking (68.7), and GLM4.1V-9B-Thinking (65.7).
Among domain-specialized models, \ours outperforms ChartCoder (73.3), RRVF (64.4), and ChartEditor (54.5) on the overall score.
These results suggest that the visual reasoning and code generation capabilities \mbox{acquired through our two-stage training generalize effectively beyond the chart editing setting.}

\paragraph{Summary.}
Across the two benchmarks, our 8B model establishes the best chart-editing performance among open-source models of comparable scale and remains competitive on direct generation,
demonstrating that \ours handles both editing and direct generation effectively.

\subsection{Ablation Study}
\label{sec:ablations}

We conduct ablation studies to validate the effect of SFT, reinforcement learning, budgeted rollout training, and the efficiency reward, with results reported on ChartEdit and ChartMIMIC benchmarks.

\begin{table*}[t]
\centering
\caption{Ablation studies on training and reward design. We start from the base model, add SFT, then apply base GRPO, budgeted rollout training, and the efficiency reward.}
\label{tab:training_reward_ablation}
\footnotesize
\setlength{\tabcolsep}{6pt}
\renewcommand{\arraystretch}{1.08}
\begin{tabular*}{\linewidth}{@{\extracolsep{\fill}}lrrrrrr@{}}
\toprule
\multirow{2}{*}{Variant} &
\multicolumn{2}{c}{ChartEdit} &
\multicolumn{2}{c}{ChartMIMIC \scriptsize {Customized}} &
\multicolumn{2}{c}{ChartMIMIC \scriptsize {Direct}} \\
\cmidrule(lr){2-3} \cmidrule(lr){4-5} \cmidrule(lr){6-7}
& Exec. & Overall & Exec. & Overall & Exec. & Overall \\
\midrule
Base & 79.15 & 67.30 & 77.61 & 61.50 & 75.00 & 58.40 \\
SFT & 87.76 & 73.40 & 80.78 & 63.90 & 84.06 & 67.80 \\
+ Base GRPO & 95.66 & 79.60 & \textbf{89.61} & \textbf{70.66} & 94.94 & 77.59 \\
+ Budgeted GRPO & 95.87 & \textbf{80.24} & 86.67 & 68.76 & \textbf{96.00} & \textbf{78.96} \\
+ Efficiency Reward & \textbf{96.65} & 79.71 & 88.61 & 70.09 & 94.72 & 77.15 \\
\bottomrule
\end{tabular*}
\end{table*}

\textbf{Training and Reward Ablation.}
Table~\ref{tab:training_reward_ablation} shows that SFT provides a clear first-stage gain over the base model, improving the overall score by 6.10 points on ChartEdit, 2.40 points on ChartMIMIC Customized Mimic, and 9.40 points on ChartMIMIC Direct Mimic.
Reinforcement learning then provides another large performance gain over the SFT checkpoint across all three settings.
Adding base GRPO improves the overall score by 6.20 points on ChartEdit, 6.76 points on ChartMIMIC Customized Mimic, and 9.79 points on ChartMIMIC Direct Mimic.
Budgeted GRPO further improves ChartEdit and Direct Mimic performance, reaching 80.24 and 78.96 overall scores, respectively.
Adding the efficiency reward preserves most of the GRPO-level performance improvement, with overall scores of 79.71 on ChartEdit, 70.09 on Customized Mimic, and 77.15 on Direct Mimic, while enabling the largest reduction in reasoning cost as shown below.

\begin{table}[t]
\centering
\caption{Average reasoning token usage under a maximum thinking budget of 16,384 tokens, starting from the base model and cumulatively adding SFT and RL variants.}
\label{tab:reasoning_token_ablation}
\footnotesize
\setlength{\tabcolsep}{7pt}
\renewcommand{\arraystretch}{1.08}
\begin{tabular*}{0.95\linewidth}{@{\extracolsep{\fill}}lrrr@{}}
\toprule
Variant & ChartEdit & ChartMIMIC Direct & ChartMIMIC Customized \\
\midrule
Base & 6521.12 & 9575.09 & 8889.43 \\
+ SFT & 2790.66 & 2267.25 & 2251.27 \\
+ Base GRPO & 3184.46 & 2320.49 & 2285.46 \\
+ Budgeted GRPO & 3040.28 & 1997.75 & 1995.39 \\
+ Efficiency Reward & \textbf{2005.18} & \textbf{1727.69} & \textbf{1515.22} \\
\bottomrule
\end{tabular*}
\end{table}

\textbf{Efficiency Reward Ablation.}
Table~\ref{tab:reasoning_token_ablation} reports average reasoning token usage under a maximum thinking budget of 16,384 tokens.
Starting from the base model, SFT alone already reduces average reasoning length from 8,328.55 to 2,436.39 tokens, a 70.8\% reduction, showing that supervised trajectory construction improves reasoning efficiency before RL is applied.
The efficiency reward further reduces token consumption across all three settings, lowering the average to 1,749.36 tokens, a 79.0\% reduction relative to the base model.
Compared with base GRPO, adding the efficiency reward reduces reasoning tokens by 37.0\% on ChartEdit, 25.5\% on Direct Mimic, and 33.7\% on Customized Mimic.
These results show that $R_{\text{eff}}$ improves the efficiency-accuracy trade-off: it retains the large performance gains brought by GRPO while suppressing unnecessary late-stage reasoning.

\section{Conclusion}
\label{sec:conclusion}

In this work, we aim to improve LRMs toward both higher editing performance and better reasoning efficiency on chart editing.
Our experiments show that reasoning indeed helps chart editing, but excessive reasoning leads to severe overthinking that ultimately degrades performance without control. 
Motivated by this, we propose \ours, a two-stage framework that supervises both the reasoning trajectory and its terminal output. 
We synthesize 200k Reason-Score-Refine trajectories via a multi-agent pipeline for SFT, and then optimize the model with reinforcement learning under a hybrid reward combining multi-dimensional fidelity reward over code, structure, and visual faithfulness with an episode-level efficiency reward. 
On chart editing benchmarks, REChart achieves state-of-the-art performance among open-source models at a comparable scale while reducing reasoning cost by over 60\% compared to the base model.

\paragraph{Limitations.}
Due to computational constraints, we fine-tune only an 8B base model, and scaling our framework to larger LRMs may yield further gains in both fidelity and reasoning efficiency. In addition, our study focuses exclusively on chart editing; extending the framework to broader visual code generation tasks such as HTML and SVG illustrations is a promising direction we leave to future work.

\subsection*{AI use statement}
LLMs serve as integral components in our data construction pipeline, including chart filtering, code synthesis, editing instruction generation, and multi-agent reasoning trajectory synthesis (Sections \ref{sec:instruction_corpus_construction} and \ref{sec:trajectory_synthesis}); the specific models assigned to each role (Qwen3-VL-32B, Qwen3-VL-235B-A22B, Gemini 3 Flash) are detailed in the corresponding sections and the Appendix. Beyond these methodological uses, LLMs were also used solely for grammar correction and language polishing of the manuscript, with no involvement in ideation, experimental design, or derivation of the scientific contributions. We have reviewed all AI-assisted work and take responsibility for the final content of this work, including text, claims, code, data, and artifacts produced with the aid of generative AI.

\subsection*{Ethics statement}
We acknowledge the broader ethical implications of generative AI. Our chart corpus is built exclusively from CC BY 4.0 licensed arXiv articles, ensuring compliance with content licensing requirements. The dataset contains no personally identifiable information (PII) and has been screened to exclude offensive content. For the human audit conducted during data filtering (Section \ref{sec:instruction_corpus_construction}), all annotators participated voluntarily with informed consent and were compensated fairly; the audit involved only assessing visual fidelity and code-image consistency of synthesized chart-code pairs, posing no foreseeable risks to participants. The release of REChart and its associated dataset poses no foreseeable harm to society and does not facilitate the generation of malicious content. All base models used in our pipeline are publicly available, and we will release our code, data, and model checkpoints upon publication.

\subsection*{Reproducibility statement}
To facilitate reproducibility, we provide a representative subset of the dataset, source code, and scripts in the supplemental material, with the full 200k SFT trajectory dataset and model checkpoints to be released on Hugging Face upon publication. Detailed experimental configurations, including agent prompts, reward formulations, training hyperparameters, and evaluation protocols, are provided in the Appendix.

\bibliographystyle{iclr2027_conference}
\bibliography{reference}

\newpage
\appendix

\section{Experimental Details of the Preliminary Study}
\label{app:prelim_study_details}

\subsection{Model Configuration}
\label{app:prelim_model_config}
We follow the standard configuration adopted in prior work~\citet{wang2023far, shi2024thorough} when querying the evaluated models.
For the open-source models in our suite (the Qwen3-VL series and GLM-4.1v-9B), we set the sampling temperature to $0.1$ and serve each model through the latest release of vLLM on a single NVIDIA A800 (80GB) GPU.
For the proprietary Gemini 3.1 Pro, we set the temperature to $0$ and query it via its official API endpoint.
All other decoding parameters are left at their respective defaults.

\subsection{Thinking Budget Implementation}
\label{app:prelim_thinking_budget}

\begin{figure*}[h]
\begin{tcolorbox}[colback=white, colframe=black, title=Prompt for Stop Reasoning]

Time is up.

I need to provide the final answer based on the existing considerations.

\end{tcolorbox}
\caption{Prompt for terminating the reasoning phase within the given thinking budget and generating the final output.}
\label{fig:end_thinking_prompt}
\end{figure*}

We implement the thinking budget through a simple two-stage decoding scheme commonly used in prior work~\citep{qwen3, qu2025optimizing}.
In the first stage, the model is allowed to reason freely with the maximum generation length set to the target thinking budget $B$.
Once generation reaches $B$ tokens, decoding is interrupted, and we append (i) an end-of-thinking prompt that instructs the model to stop reasoning and produce the final code (Figure~\ref{fig:end_thinking_prompt}) and (ii) the special end-of-thinking token \texttt{</think>}.
In the second stage, the model continues decoding from this augmented context with a fixed maximum generation length of $4{,}096$ tokens, and its output is taken as the final answer (i.e., the generated code).
For the no-thinking baseline, we skip the first stage entirely and directly prompt the model with the end-of-thinking marker, so that decoding begins in the answer phase.

\subsection{Evaluation Method}
\label{app:prelim_evaluation}
We adopt the official evaluation method of the ChartEdit benchmark~\citep{zhao2025chartedit}.
Each generated script is executed in a sandboxed Python environment, and its rendered output together with the source code is then scored by GPT-4o~\citep{gpt4o} under the official ChartEdit evaluation prompt, which assesses correctness along multiple dimensions covering both code fidelity and visual fidelity, following ChartEdit~\citep{zhao2025chartedit}.
Samples whose generated code fails to execute are assigned a score of zero on all dimensions.
The final score reported for each model and thinking budget is the average across all evaluation samples.

\subsection{Additional Reasoning Scaling Analysis}
\label{app:additional_reasoning_scaling}

\begin{figure*}[ht]
\centering
\begin{subfigure}[t]{0.32\textwidth}
    \centering
    \includegraphics[width=\linewidth]{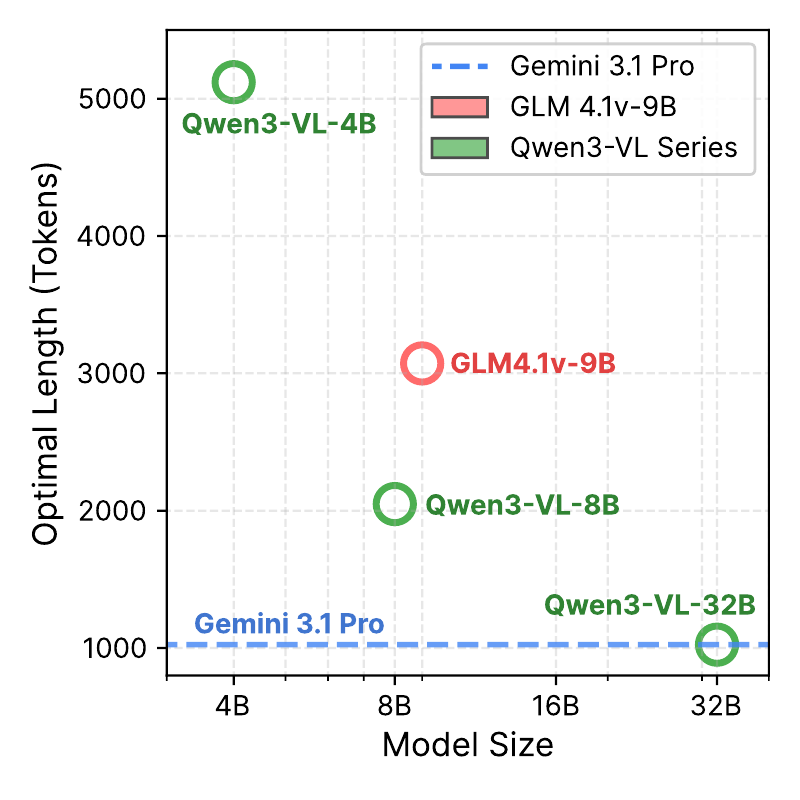}
    \caption{Optimal thinking budget across models of different scale and capability.}
    \label{fig:app_size_vs_optimal_len}
\end{subfigure}%
\hspace{0.025\textwidth}
\begin{subfigure}[t]{0.33\textwidth}
    \centering
    \includegraphics[width=\linewidth]{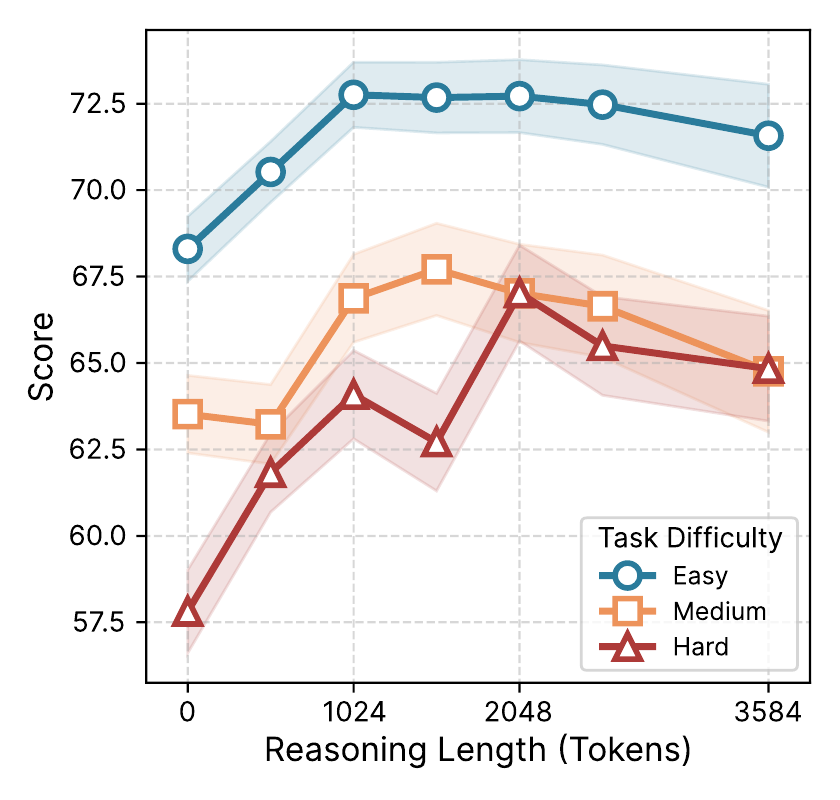}
    \caption{Qwen3-VL-8B performance under different thinking budgets and task difficulties.}
    \label{fig:app_thk_budget_vs_acc}
\end{subfigure}
\caption{
Additional analysis of reasoning budget sensitivity. (a) The optimal thinking budget differs across model scales and capabilities, suggesting that stronger models can often compress useful reasoning into shorter chains. (b) For Qwen3-VL-8B, the preferred budget also varies with task difficulty: harder editing tasks benefit from longer reasoning before overthinking begins.
}
\label{fig:app_reasoning_scaling}
\end{figure*}

Beyond the main preliminary analysis in Section~\ref{sec:preliminary_study}, we further examine two factors that modulate the effective reasoning budget.
Figure~\ref{fig:app_size_vs_optimal_len} compares the optimal thinking budget across models and shows that the best budget is not fixed across model families or scales.
Figure~\ref{fig:app_thk_budget_vs_acc} analyzes Qwen3-VL-8B across task difficulties and shows that harder examples can benefit from longer reasoning, although performance still declines once the model exceeds a useful budget.
These observations support our choice to train models toward budget-aware and contribution-aware reasoning rather than simply encouraging longer chains of thought.

\subsection{Training-Induced Reasoning Error Shift}
\label{app:training_error_shift}

\begin{table}[ht]
\centering
\caption{Reasoning error counts on the ChartEdit human set before and after training.}
\label{tab:training_error_shift}
\small
\setlength{\tabcolsep}{7pt}
\begin{tabular}{lccc}
\toprule
Variant & Visual Hallucination & Edit Intent Drift & Details Missing \\
\midrule
Base & 125 & 16 & 73 \\
+ SFT & 105 & 13 & \textbf{66} \\
+ RL & \textbf{84} & \textbf{9} & 69 \\
\bottomrule
\end{tabular}
\end{table}

SFT already reduces all three error types relative to the base model.
RL further lowers hallucination and intent drift, while slightly increasing details-missing errors relative to SFT.
This suggests that later-stage training mainly suppresses spurious reasoning rather than uniformly improving every error category.

\section{Dataset Construction Details}
\label{app:data_construction_details}

\textbf{(1) Chart Collection and Filtering.} 

To cover the diversity of chart types and usage scenarios in real-world charts, we begin with roughly 244k arXiv papers published in 2024, of which around 100k carry a CC BY 4.0 license and form our source corpus.
From their LaTeX sources, we extract about 854k figures spanning PNG, JPG, JPEG, and PDF formats.
PDF figures are rasterized to PNG, and the remaining formats are normalized to PNG so that every input shares a uniform modality.
We then apply Qwen3-VL-Flash to classify each figure as chart or non-chart, leaving about 622k chart images for downstream processing.

\textbf{(2) Chart-Code Pair Generation.} 

\begin{figure*}[ht]
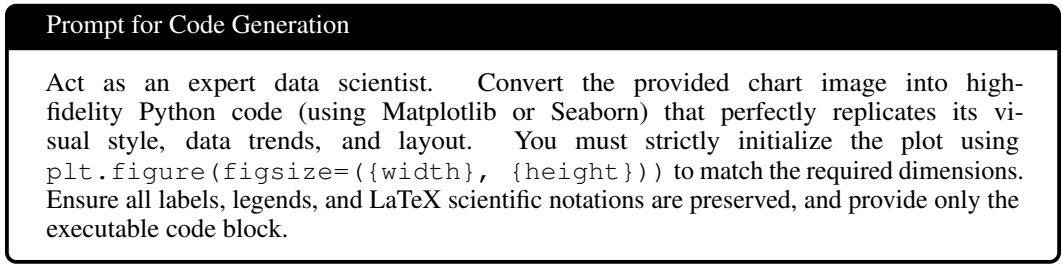

\begin{tcolorbox}[colback=white, colframe=black, title=Prompt for Code Generation]

Act as an expert data scientist. Convert the provided chart image into high-fidelity Python code (using Matplotlib or Seaborn) that perfectly replicates its visual style, data trends, and layout. You must strictly initialize the plot using \verb+plt.figure(figsize=({width}, {height}))+ to match the required dimensions. Ensure all labels, legends, and LaTeX scientific notations are preserved, and provide only the executable code block.

\end{tcolorbox}
\caption{Prompt for code reproduction based on the given image from the arXiv papers.}
\label{fig:code_regen_prompt}
\end{figure*}

\begin{figure*}[ht]
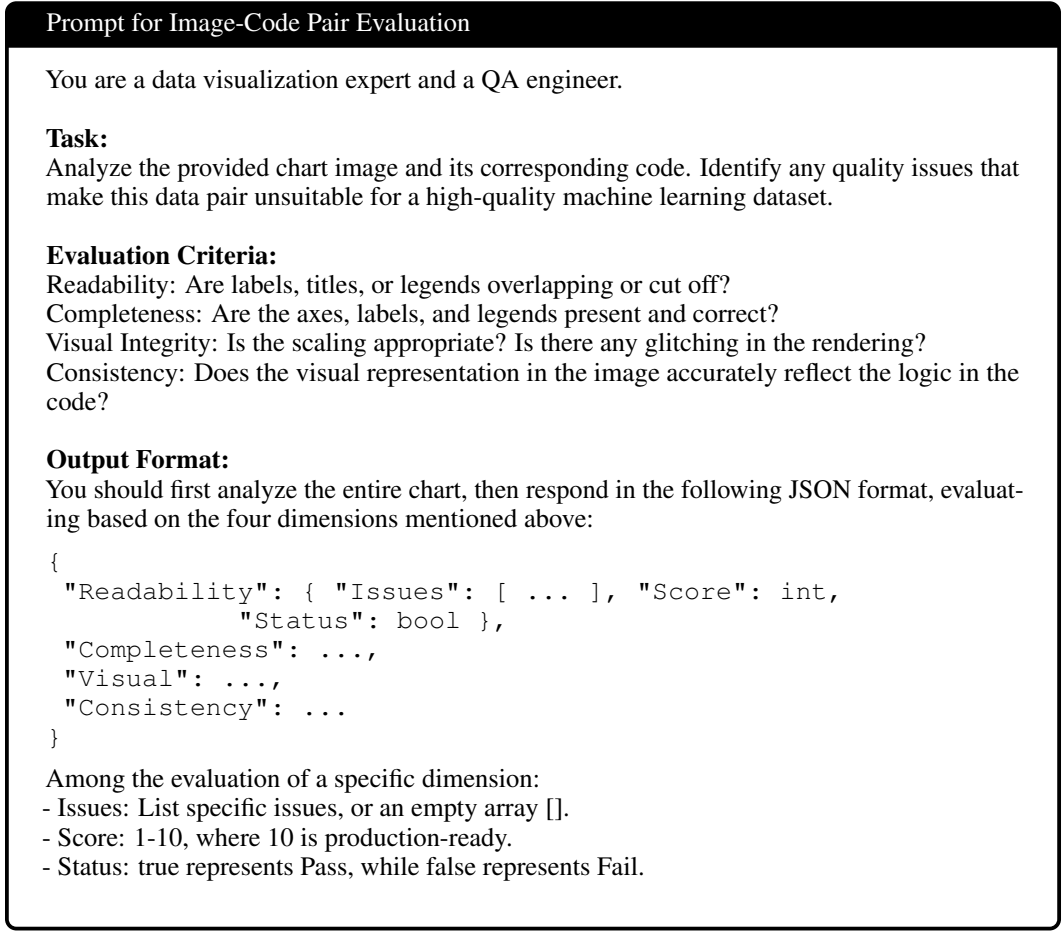

\begin{tcolorbox}[colback=white, colframe=black, title=Prompt for Image-Code Pair Evaluation]

You are a data visualization expert and a QA engineer.\\

\textbf{Task:}\\
Analyze the provided chart image and its corresponding code. Identify any quality issues that make this data pair unsuitable for a high-quality machine learning dataset.\\

\textbf{Evaluation Criteria:}\\
Readability: Are labels, titles, or legends overlapping or cut off?\\
Completeness: Are the axes, labels, and legends present and correct?\\
Visual Integrity: Is the scaling appropriate? Is there any glitching in the rendering?\\
Consistency: Does the visual representation in the image accurately reflect the logic in the code?\\

\textbf{Output Format:}\\
You should first analyze the entire chart, then respond in the following JSON format, evaluating based on the four dimensions mentioned above:

\begin{verbatim}
{
	"Readability": { "Issues": [ ... ], "Score": int,
            "Status": bool },
	"Completeness": ...,
	"Visual": ...,
	"Consistency": ... 
}
\end{verbatim}

Among the evaluation of a specific dimension:\\
- Issues: List specific issues, or an empty array [].\\
- Score: 1-10, where 10 is production-ready.\\
- Status: true represents Pass, while false represents Fail.\\

\end{tcolorbox}
\caption{Prompt for evaluating chart-code pairs during dataset construction. The prompt instructs the MLLM (Multimodal Large Language Model) assessor to check quality across four key dimensions: readability, completeness, visual integrity, and consistency. Each dimension is scored 1-10, with a binary pass/fail status. This quality filtering ensures the dataset contains only high-fidelity image-code pairs.}
\label{fig:code_regen_eval_prompt}
\end{figure*}

To generate image-code pairs at scale, we adopt a three-pass cascaded strategy that progressively assigns stronger generators to harder cases, balancing cost and quality.
In each pass, a generator is prompted with the chart image (full prompt in Figure~\ref{fig:code_regen_prompt}) to produce visualization code.
The rendered output is then evaluated against the reference chart by a Qwen3-VL-32B-Instruct judge along four dimensions (readability, completeness, visual integrity, and consistency), following the rubric in Figure~\ref{fig:code_regen_eval_prompt}.
Charts whose rendered code passes the judge are retained, while those that fail are forwarded to the next pass with a stronger generator.
We started with Qwen3-VL-32B-Instruct, then handed over the failed cases to Qwen3-VL-235B-A22B. The most challenging cases that are still unresolved are then forwarded to Gemini 3 Flash.
This process yields around 521k high-quality chart-code pairs in total.

\textbf{(3) Chart Editing Augmentation.}

To extend the dataset to chart editing, we augment each chart-code pair through an editing-instruction synthesis pipeline driven by Qwen3-VL-32B-Instruct.
For each pair, the model is prompted (Figure~\ref{fig:edit_qa_gen_prompt}) to select an appropriate editing operation from the five editing categories defined in Section \ref{sec:task_definition},  compose a natural-language editing instruction, and produce the corresponding edited code.
We repeat this process three times per pair to encourage operation and instruction diversity.
Each candidate edit is then evaluated against the criteria in  Figure~\ref{fig:edit_qa_eval_prompt}, which rates the rendered edit on four dimensions (alignment with the instruction, integrity of unmodified regions, linguistic quality, and visual legibility),
and is accepted only if it meets the required standards.

This step yields about 567k accepted editing samples.
The remaining roughly 142k chart-code pairs, for which no editing candidate is accepted, are repurposed as direct-reproduction samples, where the instruction asks the model to reconstruct the reference chart and the ground-truth code generates that reference.
The final dataset comprises around 700k image-instruction-code triplets spanning the six functional dimensions of Table~\ref{tab:task_definition}, covering both direct generation and editing tasks.
Detailed per-stage statistics, including the per-model contribution to the cascade and the distribution across editing categories, are reported in Appendix~\ref{app:dataset_statistics}.

\section{Dataset Statistic}
\label{app:dataset_statistics}


\begin{table}[ht]
\centering
\caption{Distribution of chart types in the Chart Edit dataset (by percentage).}
\label{tab:our_full_set_chart_types}
\begin{tabular}{l r l r}
\toprule
\textbf{Chart Type} & \textbf{Ratio (\%)} & \textbf{Chart Type} & \textbf{Ratio (\%)} \\
\midrule
line        & 39.74    & area        & 4.70 \\
scatters    & 14.97    & graph       & 3.52 \\
bar         & 8.70     & contour     & 3.19 \\
heatmap     & 8.50     & hist        & 3.15 \\
errorbar    & 6.99     & density     & 2.12 \\
box         & 1.43     & quiver      & 0.77 \\
scatter     & 0.66     & violin      & 0.43 \\
radar       & 0.37     & pie         & 0.28 \\
surface     & 0.17     & other       & 0.25 \\
errorpoint  & 0.08     &             &        \\
\bottomrule
\end{tabular}
\end{table}

\begin{table}[ht]
\centering
\caption{Statistics of visualization code generation models and their respective contribution to the final dataset. The table shows the number of chart-code pairs generated by each model and their proportion in the total dataset (N=521,338).}
\label{tab:codegen_models}
\begin{tabular}{lcc}
\toprule
\textbf{Model} & \textbf{Number of Charts} & \textbf{Proportion} \\
\midrule
Qwen3-VL-32B-Instruct & 118,577 & 22.74\% \\
Qwen3-VL-253B-A22B & 187,907 & 36.04\% \\
Gemini 3 Flash & 214,854 & 41.21\% \\
\midrule
\textbf{Total} & \textbf{521,338} & \textbf{100\%} \\
\bottomrule
\end{tabular}
\end{table}

\begin{table}[htbp]
\centering
\caption{Distribution of edit types in the Chart Edit dataset (by percentage).}
\label{tab:edit_type_distribution}
\begin{tabular}{l r l r}
\toprule
\textbf{Task Type} & \textbf{Ratio (\%)} & \textbf{Task} & \textbf{Ratio (\%)} \\
\midrule
Axes \& Annotations & 27.11 & Layout \& Spatial Organization & 10.02 \\
Content Manipulation \& Data Logic & 20.68 & Style \& Aesthetics & 12.68 \\
Chart Transformation & 11.61 & direct & 17.91 \\
\bottomrule
\end{tabular}
\end{table}

\begin{table}[htbp]
\centering
\caption{Pipeline statistics for image-instruction-code triplet construction.
Each row reports the count of items remaining after the corresponding stage.}
\label{tab:dataset_construction_pipeline_statistic}
\begin{tabular}{lr}
\toprule
\textbf{Stage} & \textbf{Count} \\
\midrule
arXiv papers (2024) & 244{,}029 \\
Papers under CC BY 4.0 license & 100{,}180 \\
Figures extracted from \LaTeX{} sources & 854{,}484 \\
Chart images after classification & 622{,}672 \\
Chart-code pairs after cascade filtering & 521{,}338 \\
\midrule
Editing samples accepted by audit & 567{,}049 \\
Direct-reproduction samples (no edit accepted) & 142{,}735 \\
\midrule
\textbf{Total image-instruction-code triplets} & \textbf{709{,}784} \\
\bottomrule
\end{tabular}
\end{table}

\begin{figure*}[ht]
\begin{tcolorbox}[colback=white, colframe=black, title=Prompt for Chart Editing Instruction Generation]

You are an expert Data Visualization Engineer and AI Training Specialist. Your task is to generate high-quality "Instruction-to-Code" pairs for fine-tuning a Multimodal LLM.\\

\noindent\textbf{Task}\\
Analyze the provided image and code, then generate a natural language instruction that either requests a single atomic modification or a composite request involving 2-3 interrelated modifications. The changes must be logically synergistic (e.g., changing the data density while simultaneously adjusting the axis scale to fit the new range).\\

\noindent\textbf{Task-Specific Guidelines: }\\
Choose some task from these categories:\\
- Content Manipulation \& Data Logic: Focus on the underlying narrative of the data. This involves statistical abstraction (e.g., adding mean/median lines), data density control (e.g., sampling high-frequency signals or binning continuous data), or re-filtering data series to emphasize specific trends, outliers, or comparisons.\\
- Style \& Aesthetics: Focus on the 'Data-to-Ink' ratio and aesthetic appeal. Adjust non-data components such as color mapping (sequential, diverging, or categorical), transparency (alpha) for handling overplotting, grid line visibility, background themes (clean vs. decorative), and the refined properties of data markers (edge colors, shapes, and sizes).\\
- Layout \& Spatial Organization: Focus on the topology of the canvas. This includes the strategic placement of legends and color bars to avoid occlusion, adjusting the figure's aspect ratio for better readability, managing subplot spacing (padding/margins), and re-arranging multiple panels into a more logical grid or combined structure.\\
- Chart Transformation (Structural Change): Focus on changing the fundamental visual metaphor. This involves switching between geometric marks (e.g., from Line to Area charts), re-configuring data alignment (e.g., from Grouped Bar to Stacked or 100\% Stacked Bar), or performing coordinate system transformations (e.g., Cartesian to Polar or Linear to Logarithmic).\\
- Axes \& Annotations (Contextual Layers): Focus on precision and semantic guidance. This includes fine-tuning axis limits and tick granularity, adding localized text callouts to explain anomalies, drawing directional arrows to guide the viewer's eye, or highlighting specific threshold regions and 'safe zones' with shaded areas.\\

\noindent\textbf{Other Requirements: }\\
- \textbf{Tone Strategy}: Use a \textbf{varied} tone. Synthesize a natural, human-like request.\\
- \textbf{Code Logic}: The modified code must be complete, executable, and preserve the original data structure unless the instruction explicitly asks for a data update.\\

\noindent\textbf{Output Format (JSON)}\\
1. \textbf{Visual \& Code Analysis}: Briefly describe the chart and identify the code segment to be modified.\\
2. \textbf{Metadata \& Instruction} (in \texttt{json}):
\begin{verbatim}
{
	"edit_categories": [ "Category name" ],
	"instruction": "Modification instruction",
	"change_log": [ "Detailed list of what was modified" ],
}
\end{verbatim}
3. \textbf{Refined Code} (in \texttt{python}):\\
\texttt{\# Fully functional code reflecting the changes}

\end{tcolorbox}
\caption{Prompt for generating chart editing instructions, which defines the role, inputs, task, domain-specific categories, and structured output format. This prompt is used to guide a multimodal LLM in producing high-quality instruction-code pairs for fine-tuning.}
\label{fig:edit_qa_gen_prompt}
\end{figure*}

\begin{figure*}[ht]
\begin{tcolorbox}[colback=white, colframe=black, title=System Prompt for Visual Editing Quality Audit]

You are a Senior Data Visualization Auditor and Quality Control Expert for AI training datasets. Your task is to evaluate the quality of a "Chart Edit" pair consisting of an instruction and the resulting visual transformation.\\

\noindent\textbf{Evaluation Criteria}\\
Please rate the following dimensions on a scale of 1 to 25 (1 = Poor, 25 = Excellent):

1. \textbf{Alignment (Faithfulness)}: Does the Modified Image accurately reflect the Edit Instruction?\\
\hspace*{2em}- 25: Perfectly executed as requested. \\
\hspace*{2em}- 1: The change is unrelated to the instruction or factually wrong.

2. \textbf{Integrity (Minimal Disturbance)}: Were non-requested elements preserved?\\
\hspace*{2em}- 25: Only the requested parts were changed; all other data and styles remain consistent.\\
\hspace*{2em}- 1: Significant, unintended changes occurred (e.g., data points shifted, titles vanished).

3. \textbf{Linguistic Quality}: Is the instruction natural, clear, and human-like?\\
\hspace*{2em}- 25: Sounds like a real user; professional and clear.\\
\hspace*{2em}- 1: Robotic, repetitive, or contains logical contradictions.

4. \textbf{Visual Legibility}: Is the resulting chart professional and readable?\\
\hspace*{2em}- 25: Clear, no overlapping text, appropriate contrast, and professional aesthetics.\\
\hspace*{2em}- 1: Broken layout, overlapping elements, or illegible fonts.\\

\noindent\textbf{Output Format (Strict JSON)}\\
Analyze the visual differences between the two images first, then provide the scores and brief comments in \texttt{json} format.

\begin{verbatim}
{
	"alignment": { "comment": ${comment}, "score": ${subscore} },
	"integrity": { "comment": ${comment}, "score": ${subscore} },
	"linguistic": { "comment": ${comment}, "score": ${subscore} },
	"legibility": { "comment": ${comment}, "score": ${subscore} },
}
\end{verbatim}

\end{tcolorbox}
\caption{System prompt for the quality audit of a visual editing process. The prompt defines the auditor's role, required inputs, evaluation criteria, with a detailed 1-25 scoring scale, and a strict JSON output format. The auditor must assess faithfulness, integrity, linguistic quality, and visual legibility of the edit.}
\label{fig:edit_qa_eval_prompt}
\end{figure*}

\subsection{Multi-Agent System Configuration}
\label{app:agent_prompts}

Table~\ref{tab:agent_backbones} lists the backbone model assigned to each agent in the trajectory synthesis pipeline.
The Reasoner and Generator agents share the backbone we use for downstream SFT (Qwen3-VL-8B-Thinking), so that the synthesized reasoning episodes lie within the distribution the student model is best able to internalize during fine-tuning.
The remaining three agents serve only as data-construction supervisors and are powered by the stronger Qwen3-VL-32B-Instruct, which provides more reliable planning, critique, and trajectory finalization.

\begin{table}[ht]
\centering
\caption{Backbone model assignment for each agent in the trajectory synthesis pipeline.}
\label{tab:agent_backbones}
\begin{tabular}{lll}
\toprule
\textbf{Agent} & \textbf{Symbol} & \textbf{Backbone} \\
\midrule
Planner    & $\mathcal{P}$ & Qwen3-VL-32B-Instruct \\
Reasoner   & $\mathcal{T}$ & Qwen3-VL-8B-Thinking  \\
Generator  & $\mathcal{G}$ & Qwen3-VL-8B-Thinking  \\
Critic     & $\mathcal{C}$ & Qwen3-VL-32B-Instruct \\
Finalizer  & $\mathcal{F}$ & Qwen3-VL-32B-Instruct \\
\bottomrule
\end{tabular}
\end{table}

\section{SFT Trajectory Synthesis Analysis}
\label{app:sft_synthesis_analysis}

\subsection{Trajectory Component Ablation}
\label{app:trajectory_component_ablation}

\begin{table*}[t]
\centering
\caption{Ablation of the Reason-Score-Refine synthesis pipeline. We evaluate whether the planner and critic are both needed for effective trajectory construction.}
\label{tab:trajectory_component_ablation}
\small
\setlength{\tabcolsep}{7pt}
\renewcommand{\arraystretch}{1.08}
\begin{tabular*}{\linewidth}{@{\extracolsep{\fill}}lrrrrr@{}}
\toprule
Planner & Critic & Pass@3 & Pass@6 & Pass@9 & Avg. Token \\
\midrule
\checkmark & \checkmark & \textbf{24.75\%} & \textbf{35.17\%} & \textbf{42.58\%} & \textbf{2086.33} \\
\phantom{\checkmark} & \checkmark & 24.00\% & 35.75\% & 42.58\% & 2416.21 \\
\checkmark & \phantom{\checkmark} & 16.58\% & 22.00\% & 24.75\% & 2200.22 \\
\phantom{\checkmark} & \phantom{\checkmark} & 15.33\% & 19.75\% & 23.42\% & 2300.86 \\
\bottomrule
\end{tabular*}
\end{table*}

The planner and critic are complementary.
Keeping both modules yields the strongest pass@k and the lowest token cost, while removing either module hurts trajectory quality.
In particular, the planner is most important for reaching high pass@k, whereas the critic mainly helps the trajectory recover from unproductive reasoning.

\subsection{Synthesis Cost}
\label{app:synthesis_cost}

\begin{table}[ht]
\centering
\caption{Estimated throughput and compute footprint of the SFT synthesis pipeline.}
\label{tab:sft_synthesis_cost}
\small
\setlength{\tabcolsep}{6pt}
\begin{tabularx}{\linewidth}{l X r}
\toprule
Module & Hardware & Throughput \\
\midrule
Planner & 1$\times$A800 & 3500/h \\
Reasoner & 1$\times$H100 & 4000/h \\
Generator & 1$\times$H100 & 4000/h \\
Scorer & 64 Core GPU + 512GB Memory & 9000/h \\
Critic & 1$\times$A800 & 18000/h \\
Finalizer & 1$\times$A800 & 9000/h \\
\midrule
\textbf{Total} & Running on two machines (8$\times$A800+64C CPU + 512GB Mem; 8$\times$H100+64C CPU + 512GB Mem) & About 2 weeks (380 hours) \\
\bottomrule
\end{tabularx}
\end{table}

The synthesis pipeline is compute-intensive but manageable with two coordinated machines.
The planner, critic, and finalizer run on A800-class hardware, while the reasoner and generator occupy H100-class GPUs, and the scorer dominates the throughput budget.

\subsection{SFT Filtering Statistics}
\label{app:sft_filtering_stats}

\begin{table}[ht]
\centering
\caption{Statistics of the SFT data synthesis pipeline after filtering.}
\label{tab:sft_filtering_stats}
\small
\setlength{\tabcolsep}{6pt}
\begin{tabular}{lrrrrr}
\toprule
Task Type & Total Input & Filtered & Final Output & Pass Rate & Avg. Round Num \\
\midrule
Edit Task & 562,485 & 383,867 & 178,618 & 31.76\% & 5.39 \\
Direct Task & 75,275 & 45,957 & 29,318 & 38.95\% & 5.42 \\
\midrule
\textbf{Total} & \textbf{637,760} & \textbf{429,824} & \textbf{207,936} & \textbf{32.60\%} & \textbf{5.40} \\
\bottomrule
\end{tabular}
\end{table}

These statistics show that the editing portion is harder to retain through the cascade, while direct-reproduction samples have a slightly higher pass rate.
Across both splits, the synthesis loop averages about 5.4 rounds before producing an accepted trajectory.

\subsection{Effect of SFT Data Scale}
\label{app:sft_data_scale}

\begin{table*}[ht]
\centering
\caption{Effect of SFT dataset size on downstream performance.}
\label{tab:sft_data_scale}
\small
\setlength{\tabcolsep}{6pt}
\renewcommand{\arraystretch}{1.05}
\begin{tabular*}{\linewidth}{@{\extracolsep{\fill}}lrrrr@{}}
\toprule
SFT Dataset Size & \multicolumn{2}{c}{ChartEdit w/o Code} & \multicolumn{2}{c}{ChartMIMIC Direct} \\
\cmidrule(lr){2-3} \cmidrule(lr){4-5}
& Exec. Rate & Overall & Exec. Rate & Overall \\
\midrule
base & 79.15 & 67.30 & 75.00 & 58.45 \\
50k & 83.56 & 68.76 & 79.06 & 61.75 \\
100k & 84.48 & 69.12 & 80.72 & 63.58 \\
150k & 86.98 & 71.01 & 82.11 & 65.05 \\
full & \textbf{87.76} & \textbf{73.43} & \textbf{84.06} & \textbf{67.78} \\
\bottomrule
\end{tabular*}
\end{table*}

The full SFT set gives the strongest downstream results on both benchmarks.
Performance improves steadily as more synthesized trajectories are added, indicating that the data scaling effect is still positive even after 150k samples, although the gains become smaller at the largest scale.

\section{Experimental Details}
\label{app:eval_config}

\subsection{Evaluation Configuration}

Following prior work~\citep{wang2023far, shi2024thorough}, we adopt the following decoding configurations for all evaluations.
For open-source models, we set the temperature to 0.1 and top-p to 1.
For proprietary models, we set the temperature to 0 and top-p to 1, with a maximum output length of 4096 tokens.
For all open-source reasoning models, we set the thinking budget to 4096 tokens and the maximum output length for final code generation to 4096 tokens; the implementation of thinking budget control follows the protocol described in Appendix~\ref{app:prelim_thinking_budget}.
All open-source models are served via the vLLM inference library on NVIDIA A800 80GB GPUs.

\subsection{Benchmark Descriptions}

\paragraph{ChartEdit~\citep{zhao2025chartedit}.}
ChartEdit is a chart editing benchmark that requires models to modify an existing chart according to a natural language instruction, without access to the original source code.
Each sample provides a reference chart image and an editing instruction, and the model is expected to generate executable visualization code that reflects the requested changes.
Performance is assessed across three complementary dimensions: code-level correctness, which measures the syntactic and structural validity of the generated code; chart-level fidelity, which evaluates the visual alignment between the generated and target charts; and execution-level success, which checks whether the code runs without errors.
An official overall score aggregates these dimensions into a single metric.

\paragraph{ChartMIMIC~\citep{yang2024chartmimic}.}
ChartMIMIC evaluates models on two distinct tasks.
The Customized Mimic task requires instruction-conditioned chart editing, where the model must adapt a reference chart according to a given instruction; samples are divided into Low and High difficulty tiers based on the complexity of the required edits.
The Direct Mimic task measures chart-to-code reproduction ability without any editing instruction, providing a complementary assessment of the model's direct chart generation capability.
Both tasks report execution success rate and an overall score computed from visual and structural similarity between the generated and reference charts.

\end{document}